\documentclass{article}
\usepackage{spconf,amsmath,epsfig}

\title{Richer Syntactic Dependencies for Structured Language Modeling}

\twoauthors
  {Ciprian Chelba}
       {Microsoft Speech.Net / Microsoft Research\\
        One Microsoft Way\\
        Redmond, WA 98052\\
        chelba@microsoft.com}
  {Peng Xu}
       {Center for Language and Speech Processing\\ 
        Johns Hopkins University\\ 
        Baltimore, MD 21218\\ 
        xp@clsp.jhu.edu}

\begin{document}
%
\maketitle
\begin{abstract}
        The paper investigates the use of richer syntactic
dependencies in the structured language model (SLM). We present two
simple methods of enriching the dependencies in the syntactic parse
trees used for intializing the SLM. We evaluate the impact of both methods
on the perplexity (PPL) and word-error-rate (WER, N-best rescoring) performance of the
SLM. We show that the new model achieves an improvement in PPL and WER
over the baseline results reported using the SLM on the
UPenn Treebank and Wall Street Journal (WSJ) corpora, respectively.
\end{abstract}
\section{Introduction}
\label{sec:intro}

The structured language model uses hidden parse trees to assign
conditional word-level language model probabilities.
As explained in~\cite{chelba00}, Section 4.4.1, the potential
reduction in PPL --- relative to a 3-gram baseline --- using the SLM's
headword parametrization for word prediction is about 40\%. The key
to achieving this is a good guess of the final best parse for a given sentence
as it is being traversed left-to-right. This is much harder
than finding the final best parse for the entire
sentence, as it is sought in a regular statistical
parser. Nevertheless, it is expected that techniques developed in the
statistical parsing community that aim at recovering the best
parse for an entire sentence, i.e.\ as judged by a human annotator, 
should be productive in reducing the PPL of the SLM as well.

In this paper we present a simple and novel way of enriching the probabilistic
dependencies in the CONSTRUCTOR component of the SLM showing that it
leads to better PPL and WER performance of the model.
Similar ways of enriching the dependency structure underlying the
parametrization of the probabilistic model used for scoring a given
parse tree are used in the statistical parsing community
\cite{charniak00},~\cite{mike:thesis}. Recently, such models
\cite{charniak01},~\cite{roark:thesis} have been
shown to outperform the SLM in terms of PPL and WER on the UPenn
Treebank and Wall Street Journal corpora, respectively. The simple
modification we present brings the WER performance of the SLM at the
same level with the best reported in~\cite{roark:thesis}, despite a
modest improvement in PPL when interpolating the SLM with a 3-gram
model.

The remaining part of the paper is organized as follows:
Section~\ref{sec:slm_overview} briefly describes the
SLM. Section~\ref{sec:hp} discusses the 
binarization and headword percolation procedure used in the standard
training of the SLM followed by a description of the procedure used
for enriching the syntactic dependencies in the SLM.
Section~\ref{sec:experiments} describes the experimental setup and
results. Section~\ref{sec:conclusions} discusses the results and
indicates future research directions.
  
\section{Structured Language Model Overview}
\label{sec:slm_overview}

An extensive presentation of the SLM can be found
in~\cite{chelba00}. The model assigns a probability $P(W,T)$ to every
sentence $W$ and its every possible binary parse $T$. The
terminals of $T$ are the words of $W$ with POStags, and the nodes of $T$ are
annotated with phrase headwords and non-terminal labels.
\begin{figure}[h]
  \begin{center}
    \epsfig{file=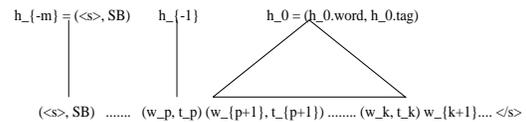,height=1.5cm,width=7cm}
  \end{center}
  \vspace{-0.5cm}
  \caption{A word-parse $k$-prefix} \label{fig:w_parse}
\end{figure}
 Let $W$ be a sentence of length $n$ words to which we have prepended
the sentence begining marker \verb+<s>+ and appended the sentence end
marker \verb+</s>+ so that $w_0 = $\verb+<s>+ and $w_{n+1} = $\verb+</s>+.
Let $W_k = w_0 \ldots w_k$ be the word $k$-prefix of the sentence ---
the words from the begining of the sentence up to the current position
 $k$ --- and  \mbox{$W_k T_k$} the \emph{word-parse $k$-prefix}. Figure~\ref{fig:w_parse} shows a
word-parse $k$-prefix; \verb|h_0 .. h_{-m}| are the \emph{exposed
 heads}, each head being a pair (headword, non-terminal label), or
(word,  POStag) in the case of a root-only tree. The exposed heads at
a given position $k$ in the input sentence are a function of the
word-parse $k$-prefix.

\subsection{Probabilistic Model} \label{ssec:prob_model}

\begin{figure}
  \begin{center} 
    \epsfig{file=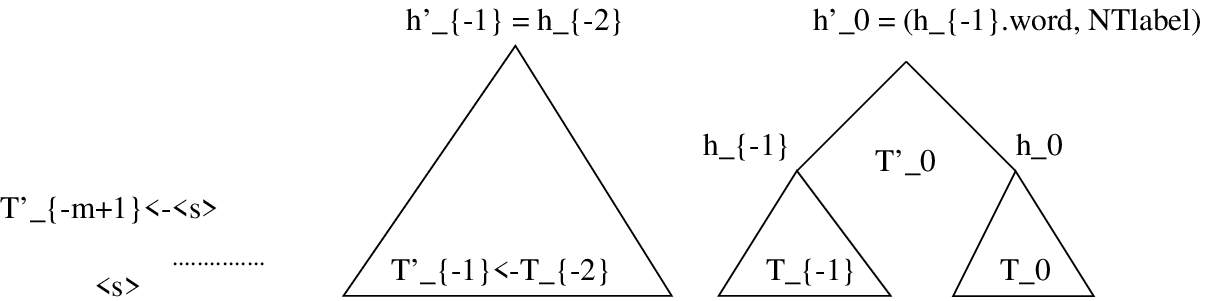,height=1.7cm,width=7cm}
  \end{center}
  \vspace{-0.5cm}
  \caption{Result of adjoin-left under NTlabel} \label{fig:after_a_l}
  \vspace{-0.5cm}
\end{figure}

\begin{figure}
  \begin{center} 
    \epsfig{file=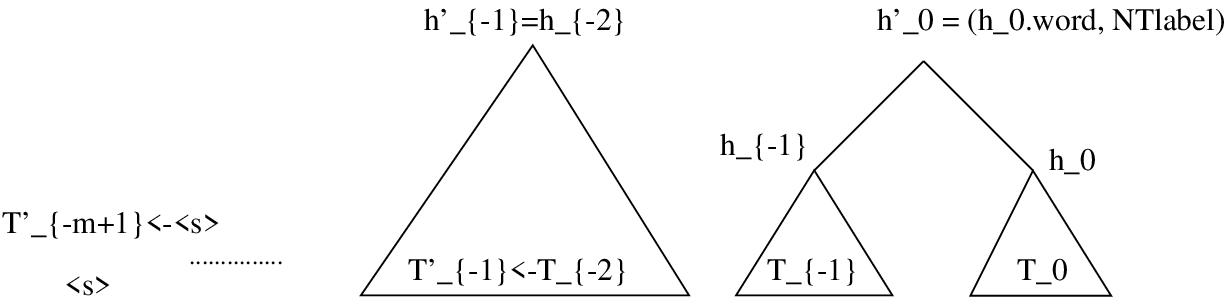,height=1.7cm,width=7cm}
  \end{center}
  \vspace{-0.5cm}
  \caption{Result of adjoin-right under NTlabel} \label{fig:after_a_r}
  \vspace{-0.5cm}
\end{figure}

 The joint probability $P(W,T)$ of a word sequence $W$ and a complete parse
$T$ can be broken into:
\begin{eqnarray}
\lefteqn{P(W,T)= } \nonumber\\
& \prod_{k=1}^{n+1}[&P(w_k/W_{k-1}T_{k-1}) \cdot P(t_k/W_{k-1}T_{k-1},w_k) \cdot \nonumber\\
&  & \prod_{i=1}^{N_k}P(p_i^k/W_{k-1}T_{k-1},w_k,t_k,p_1^k\ldots
p_{i-1}^k)] \label{eq:model}
\end{eqnarray}
where: \\
$\bullet$ $W_{k-1} T_{k-1}$ is the word-parse $(k-1)$-prefix\\
$\bullet$ $w_k$ is the word predicted by WORD-PREDICTOR\\
$\bullet$ $t_k$ is the tag assigned to $w_k$ by the TAGGER\\
$\bullet$ $N_k - 1$ is the number of operations the CONSTRUCTOR executes at 
sentence position $k$ before passing control to the  WORD-PREDICTOR
(the $N_k$-th operation at position k is the \verb+null+ transition);
$N_k$ is a function of $T$\\
$\bullet$ $p_i^k$ denotes the $i$-th CONSTRUCTOR operation carried out at
position k in the word string; the operations performed by the
CONSTRUCTOR are illustrated in
Figures~\ref{fig:after_a_l}-\ref{fig:after_a_r} and they ensure that
all possible binary branching parses with all possible headword and
non-terminal label assignments for the $w_1 \ldots w_k$ word
sequence can be generated. The $p_1^k \ldots p_{N_k}^k$ sequence of CONSTRUCTOR
operations at position $k$ grows the word-parse $(k-1)$-prefix into a
word-parse $k$-prefix.
 
Our model is based on three probabilities, each estimated using deleted
interpolation and parameterized (approximated) as follows:
\begin{eqnarray}
  P(w_k/W_{k-1} T_{k-1}) & = & P(w_k/h_0, h_{-1})\label{eq:1}\\
  P(t_k/w_k,W_{k-1} T_{k-1}) & = & P(t_k/w_k, h_0, h_{-1})\label{eq:2}\\
  P(p_i^k/W_{k}T_{k}) & = & P(p_i^k/h_0, h_{-1})\label{eq:3}
\end{eqnarray}%
 It is worth noting that if the binary branching structure
developed by the parser were always right-branching and we mapped the
POStag and non-terminal label vocabularies to a single type then our
model would be equivalent to a trigram language model.
 Since the number of parses  for a given word prefix $W_{k}$ grows
exponentially with $k$, $|\{T_{k}\}| \sim O(2^k)$, the state space of
our model is huge even for relatively short sentences, so we had to use
a search strategy that prunes it. Our choice was a synchronous
multi-stack search algorithm which is very similar to a beam search.

The \emph{language model} probability assignment for the word at position $k+1$ in the input
sentence is made using:
\begin{eqnarray}
P_{SLM}(w_{k+1}/W_{k}) & =
& \sum_{T_{k}\in
  S_{k}}P(w_{k+1}/W_{k}T_{k})\cdot\rho(W_{k},T_{k}),\nonumber \\ 
\rho(W_{k},T_{k}) & = & P(W_{k}T_{k})/\sum_{T_{k} \in S_{k}}P(W_{k}T_{k})\label{eq:ppl1}
\end{eqnarray}
which ensures a proper probability over strings $W^*$, where $S_{k}$ is
the set of all parses present in our stacks at the current stage $k$.

Each model component --- WORD-PREDICTOR, TAGGER, CONSTRUCTOR ---
is initialized from a set of parsed sentences after undergoing
headword percolation and binarization, see Section~\ref{sec:hp}. An
N-best EM~\cite{em77} variant is then employed to jointly reestimate
the model parameters such that the PPL on training data is decreased
--- the likelihood of the training data under our model is
increased. The reduction in PPL is shown experimentally to carry over
to the test data.

\section{Headword Percolation and Binarization}
\label{sec:hp}

As explained in the previous section, the SLM is initialized on parse
trees that have been binarized and the non-terminal (NT) tags at each node
have been enriched with headwords. We will briefly review the headword
percolation and binarization procedures; they are explained in
detail in~\cite{chelba00}.

The position of the headword within a constituent
--- equivalent with a context-free production of the type\\
$Z \rightarrow Y_1 \ldots Y_n$ , where $Z, Y_1, \ldots Y_n$ are
NT labels or POStags (only for $Y_i$) ---  is identified
using a rule-based approach.

\begin{figure}
  \begin{center} 
    \epsfig{file=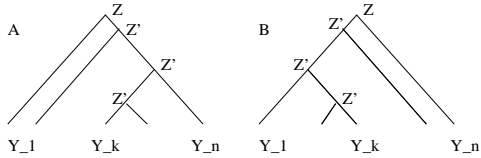,height=2cm,width=6cm}
  \end{center}
  \caption{Binarization schemes} \label{fig:bin_schemes}
\end{figure}

Assuming that the index of the headword on the right-hand side of the
rule is $k$, we binarize the constituent as follows: depending on the
$Z$ identity we apply one of the two binarization schemes in
Figure~\ref{fig:bin_schemes}. The intermediate nodes created by the
above binarization schemes receive the NT label
$Z'$\footnote{Any resemblance to X-bar theory is purely coincidental.}. 
The choice among the two schemes is made according to a list of
rules based on the identity of the label on the left-hand-side of a CF
rewrite rule.

Under the equivalence classification in Eq.~(\ref{eq:3}), the conditioning
information available to the CONSTRUCTOR model component is the
two most-recent exposed heads consisting of two NT tags and
two headwords. In an attempt to extend the syntactic dependencies
beyond this level, we enrich the non-terminal tag of a node in the
binarized parse tree with the NT tag of one if its children
or both. We distinguish between two ways of picking the child from
which the NT tag is being percolated: 
\begin{enumerate}
\item{\underline{same}}: we use the non-terminal tag of the node from which the
headword is being percolated
\item{\underline{opposite}}: we use the non-terminal tag of the sibling node from which the
headword is being percolated
\end{enumerate}

For example, the noun phrase constituent 
\begin{verbatim}
(NP 
    (DT the)
    (NNP dutch)
    (VBG publishing)
    (NN group))
\end{verbatim}
becomes
\begin{verbatim}
(NP_GROUP
    (DT the)
    (NP'_GROUP
        (NNP dutch)
        (NP'_GROUP (VBG publishing)
                   (NN group))))
\end{verbatim}
after binarization and headword percolation and
\begin{verbatim}
(NP+NP'_GROUP
    (DT the)
    (NP'+NP'_GROUP
         (NNP dutch)
         (NP'+NN_GROUP (VBG publishing)
                       (NN group))))
\end{verbatim}
or
\begin{verbatim}
(NP+DT_GROUP 
    (DT the) 
    (NP'+NNP_GROUP 
         (NNP dutch) 
         (NP'+VBG_GROUP (VBG publishing) 
                        (NN group))))
\end{verbatim}
after enriching the non-terminal tags using the \emph{same} and \emph{opposite}
scheme, respectively.

A given binarized tree is traversed recursively in depth first order and each
constituent is enriched in the above manner. The SLM is then
initialized on the resulting parse trees.

Although it is hard to find a direct correspondence between the above
way of enriching the dependency structure of the probability
model and the ones used
in~\cite{charniak00},~\cite{charniak01}~or~\cite{roark:thesis}, they
are similar.

\section{Experiments}
\label{sec:experiments}

We have evaluated the PPL performance of the model on the UPenn
Treebank and the WER performance in the setups described
in~\cite{chelba00}, respectively.

\subsection{Perplexity experiments on the UPenn Treebank}
\label{ssec:ppl}

For convenience, we chose to evaluate the performance of the enriched SLM on the UPenn
Treebank corpus~\cite{Upenn} --- a subset of the Wall Street
Journal (WSJ) corpus~\cite{wsj0}.

We have evaluated the perplexity of the two different ways of
enriching the non-terminal tags in the parse tree and of using
both of them at the same time. For each way of initializing the SLM we
have performed 3 iterations of N-best EM. The word and POS-tagger
vocabulary sizes were 10,000 and 40, respectively. The NT
tag/CONSTRUCTOR operation vocabulary sizes were 52/157, 954/2863,
712/2137, 3816/11449 for the baseline, \emph{opposite}, \emph{same} and both
initialization schemes, respectively. The SLM is interpolated with
a 3-gram model --- built on exactly the same training data/word
vocabulary as the SLM --- using a fixed interpolation weight:
$$P(\cdot)=\lambda \cdot P_{3gram}(\cdot) +(1-\lambda) \cdot
P_{SLM}(\cdot)$$
The results are summarized in Table~\ref{tab:ppl_results}. The
\emph{baseline} model is the standard SLM as described in~\cite{chelba00}.
\begin{table}[htbp]
  \begin{center}
    \begin{tabular}{|l|l|r|r|r|}\hline
      Model             & Iter                  & $\lambda$ = 0.0       & $\lambda$ = 0.6       & $\lambda$ = 1.0\\\hline
      baseline          & 0                     & 167.38                & 151.89                & 166.63\\
      baseline          & 3                     & 158.75                & 148.67                & 166.63\\ \hline
      opposite          & 0                     & 157.61                & 146.99                & 166.63\\
      opposite          & 3                     & 150.83                & \underline{144.08}    & 166.63\\ \hline
      same              & 0                     & 163.31                & 149.56                & 166.63\\
      same              & 3                     & 155.29                & 146.39                & 166.63\\ \hline
      both              & 0                     & 160.48                & 147.52                & 166.63\\
      both              & 3                     & 153.30                & 144.99                & 166.63\\ \hline
    \end{tabular}
    \caption{Deleted Interpolation 3-gram $+$ SLM; PPL Results}
    \label{tab:ppl_results}
  \end{center}
  \vspace{-0.5cm}
\end{table}
As can be seen, the model initialized using the \emph{opposite} scheme
performed best, reducing the PPL of the SLM by 5\% relative to the
SLM baseline performance. However the improvement in PPL is minor
after interpolating with the 3-gram model.

\begin{table*}
  \begin{center}
    \begin{tabular}{|l|l|r|r|r|r|r|r|}\hline
      Model                     & Iter                  &
      \multicolumn{6}{c|}{Interpolation weight}\\ \cline{3-8}
                                &                       & 0.0   & 0.2   & 0.4   & 0.6   & 0.8   & 1.0   \\ \hline
      baseline SLM WER, \%              & 0                     & 13.1  & 13.1  & 13.1  & 13.0  & 13.4  & \bf{13.7}     \\\hline
      opposite SLM WER, \%              & 0                     & 12.7  & 12.8  & 12.7  & \underline{12.7} & 13.1  & \bf{13.7}\\\hline
      \multicolumn{2}{|l|}{MPSS significance test p-value}      
                                                        & 0.020 & 0.017 & 0.014 & \underline{0.005} & 0.070  & ---\\ \hline
    \end{tabular}
    \caption{Back-off 3-gram $+$ SLM; N-best rescoring WER Results and
Statistical Significance }
    \label{tab:wer_results}
  \end{center}
  \vspace{-0.5cm}
\end{table*}

\subsection{N-best rescoring results}
\label{ssec:wer}

We chose to evaluate in the WSJ DARPA'93 HUB1 test setup. The size
of the test set is 213 utterances, 3446 words. The 20kwds open
vocabulary and baseline 3-gram model --- used for generating the
lattices and the N-best lists --- are the standard ones provided by
NIST and LDC --- see~\cite{chelba00} for details. The SLM was trained
on 20Mwds of WSJ text automatically parsed using the parser
in~\cite{ratnaparkhi:parser}, binarized and enriched with headwords
and the \emph{opposite} NT tag information as explained in
Section~\ref{sec:hp}. The results are presented in Table~\ref{tab:wer_results}.

Since the rescoring experiments are expensive, we have only evaluated
the WER performance of the model intialized using the \emph{opposite} scheme. 
The enriched SLM achieves 0.3-0.4\% absolute
reduction in WER over the performance of the baseline SLM and a full
1.0\% absolute over the baseline 3-gram model,
for a wide range of values of the interpolation weight. We note that the performance
of the SLM as a second pass language model is the same even without
interpolating it with the 3-gram model\footnote{The
N-best lists are generated using the baseline 3-gram model so this is
not indicative of the performance of the SLM as a first pass language
model.} ($\lambda = 0.0$).

We have evaluated the statistical significance of the results
relative to the 3-gram baseline
using the standard test suite in the SCLITE package provided by
NIST. We believe that for WER statistics the most relevant significance test
is the Matched Pair Sentence Segment one. The results are presented in
Table~\ref{tab:wer_results}. As it can be seen the improvement
achieved by the SLM is highly significant at all values of the
interpolation weight $\lambda$ except for $\lambda = 0.8$.

\section{Conclusions and Future Directions}
\label{sec:conclusions}

We have presented a simple but effective method of enriching the syntactic
dependencies in the structured language model (SLM) that achieves
0.3-0.4\% absolute reduction in WER over the best previous results
reported using the SLM on WSJ. The implementation could be greatly
improved by predicting only the relevant part of the enriched
non-terminal tag and then adding the part inherited from the
child. A more comprehensive study of the most productive ways of
increasing the probabilistic dependencies in the parse tree would be
desirable.

\section{Acknowledgements}
\label{sec:ack}

The authors would like to thank Brian Roark for making
available the N-best lists for the HUB1 test set. The
SLM is publicly available at:\\ftp://ftp.clsp.jhu.edu/pub/clsp/chelba/SLM\_RELEASE.

\bibliographystyle{//chelba/WORK/Documents/latex/IEEEbib}
\bibliography{//chelba/WORK/Documents/latex/mainbibfile}

\end{document}